\documentclass[letterpaper, 10 pt, conference]{ieeeconf}  % Comment this line out if you need a4paper

\IEEEoverridecommandlockouts                              % This command is only needed if 
\usepackage{graphicx} % for pdf, bitmapped graphics files
\usepackage{amsmath} % assumes amsmath package installed
\usepackage{amssymb}  % assumes amsmath package installed
\usepackage{subcaption}
\usepackage{siunitx}
\usepackage{multirow}
\usepackage{cite}
\usepackage{bm}
\usepackage{mathtools}
\usepackage{pifont}
\usepackage[dvipsnames]{xcolor}
\usepackage[font=footnotesize]{caption}
\usepackage[abbreviations]{glossaries-extra}
\usepackage{booktabs}
\usepackage[colorlinks = true,
            linkcolor = blue,
            urlcolor  = blue,
            citecolor = blue,
            anchorcolor = blue]{hyperref}
\usepackage[capitalise]{cleveref}
\usepackage[table]{xcolor}
\usepackage{makecell}

\newcommand{\approptoinn}[2]{\mathrel{\vcenter{
  \offinterlineskip\halign{\hfil$##$\cr
    #1\propto\cr\noalign{\kern2pt}#1\sim\cr\noalign{\kern-2pt}}}}}

\usepackage{float}
\usepackage{flushend}
\usepackage{comment}
\definecolor{earthgreen}{RGB}{230,255,230}
\definecolor{marsred}{RGB}{255,230,230}
\definecolor{moongrey}{RGB}{240,240,240}
\definecolor{superblue}{RGB}{210,240,255}

\title{End-to-End Reinforcement Learning for Quadrupedal Jumping \\ in Lunar Gravity}

\author{Levin Müller, Oliver Fischer, Marco Trentini, Yusuke Tanaka, and 
Marco Hutter% <-this % stops a space
\thanks{All authors are with ETH Zurich, Robotics Systems Lab; Leonhardstrasse 21, 8092 Zurich, Switzerland.
        Contact: {\tt\small muellerlev@ethz.ch}}%
\thanks{
}
}
\begin{document}

% This allows you to play around with how the references show up in the paper -- can be used to make space :))
\bstctlcite{IEEEexample:BSTcontrol}

\maketitle
\thispagestyle{empty}
\pagestyle{empty}

%%%%%%%%%%%%%%%%%%%%%%%%%%%%%%%%%%%%%%%%%%%%%%%%%%%%%%%%%%%%%%%%%%%%%%%%%%%%%%%%
\begin{abstract}
Legged robots can extend planetary exploration to rugged and discontinuous terrain that is inaccessible to conventional rovers. In lunar gravity, jumping further expands this mobility by enabling robots to cross obstacles and gaps that cannot be traversed by walking alone. We present a multistage reinforcement-learning approach that produces a single end-to-end control policy for commanded point-to-point jumping under lunar gravity. The resulting policy has a simple architecture and relies exclusively on mission-compatible onboard sensing, making its deployment on a real lunar mission plausible. In simulation, the learned policy performs accurate point-to-point jumps with landing errors smaller than the robot's body size. We analyze the torque and power demands of lunar jumping and characterize performance degradation under extreme outlier events, including control-update freezes of up to \SI{1}{s} during flight. Hardware transfer demonstrates that the policy can successfully execute a vertical jump under lunar-gravity offloading. Furthermore, the measured torque and power profiles are consistent with the simulation results and remain within the robot's hardware limits.
\end{abstract}

%%%%%%%%%%%%%%%%%%%%%%%%%%%%%%%%%%%%%%%%%%%%%%%%%%%%%%%%%%%%%%%%%%%%%%%%%%%%%%%%

\section{INTRODUCTION}

Robotic exploration of the solar system is essential for advancing scientific understanding of planetary bodies and preparing future human missions. While orbiters and stationary landers provide valuable measurements, scientifically relevant targets require mobile surface systems to acquire local observations and interact with the terrain. As interest grows in sustained surface operations~\cite{krawciw2026lunar}, infrastructure construction~\cite{uno2025moonbot, brinkmann2020enhancement}, and scientific access to challenging terrain, missions increasingly target sites such as permanently shadowed regions, crater walls, and lunar lava tubes -- areas that remain difficult to reach with conventional mobility systems~\cite{jangale2025novel, cordes2011lunares, kolvenbach2025lunarleaper, parness_lemur_2017, tanaka2025scaler, schperberg2025mobius, uno2026limbero}. Planetary rovers have so far relied predominantly on wheeled locomotion, which provides substantial flight heritage and reliability on comparatively benign terrain, but remains limited on steep, loose, rugged, or discontinuous surfaces \cite{ishigami2009slope}. In contrast, recent advances in deep reinforcement learning have substantially improved the ability of legged robots to traverse rough and unstructured terrain, motivating their investigation for planetary exploration.% and, in particular, the selection of a small legged robot for the \textit{LunarLeaper} mission concept~\cite{kolvenbach2025lunarleaper}.
\\
Beyond robust walking, the morphology of legged robots also enables dynamic locomotion modes such as jumping \cite{spiridonov2024spacehopper}. For a lunar robot, jumping is not intended to replace walking as an energy-efficient mode of long-range traversal. Still, it can serve as a useful mobility primitive for overcoming obstacles, terrain discontinuities, slopes, and gaps that would otherwise be untraversable~\cite{spacebok}. In low gravity, such maneuvers are especially attractive because relatively moderate launch velocities can produce extended ballistic flight phases \cite{spiridonov2024spacehopper}.

At the same time, lunar jumping imposes additional control and deployment challenges: the controller must demonstrate high robustness, particularly during flight and landing, to avoid crashing onto the surface. Specifically on non-redundant system architectures, FDIR may even require the ability to recover from a mid-flight system reboot. In addition, constraints in computing hardware for aerospace engineering favor the development of a low-complexity controller. Although quadrupedal jumping has been demonstrated in both terrestrial and low-gravity contexts, commanded point-to-point jumping for a quadruped in lunar gravity has not yet been sufficiently investigated.
\\
To address this gap, we develop a deployment-lean reinforcement-learning framework for commanded point-to-point jumping in lunar gravity. The key idea is to pair the governing ballistic properties of lunar flight with a multistage control scheme. This provides a versatile training framework to synthesize an end-to-end deep reinforcement learning-based jumping controller within a single multilayer perceptron control policy. The control policy maps onboard-feasible observations to joint-position targets.

\begin{figure}[!t]
    \centering
    \includegraphics[width=\columnwidth]{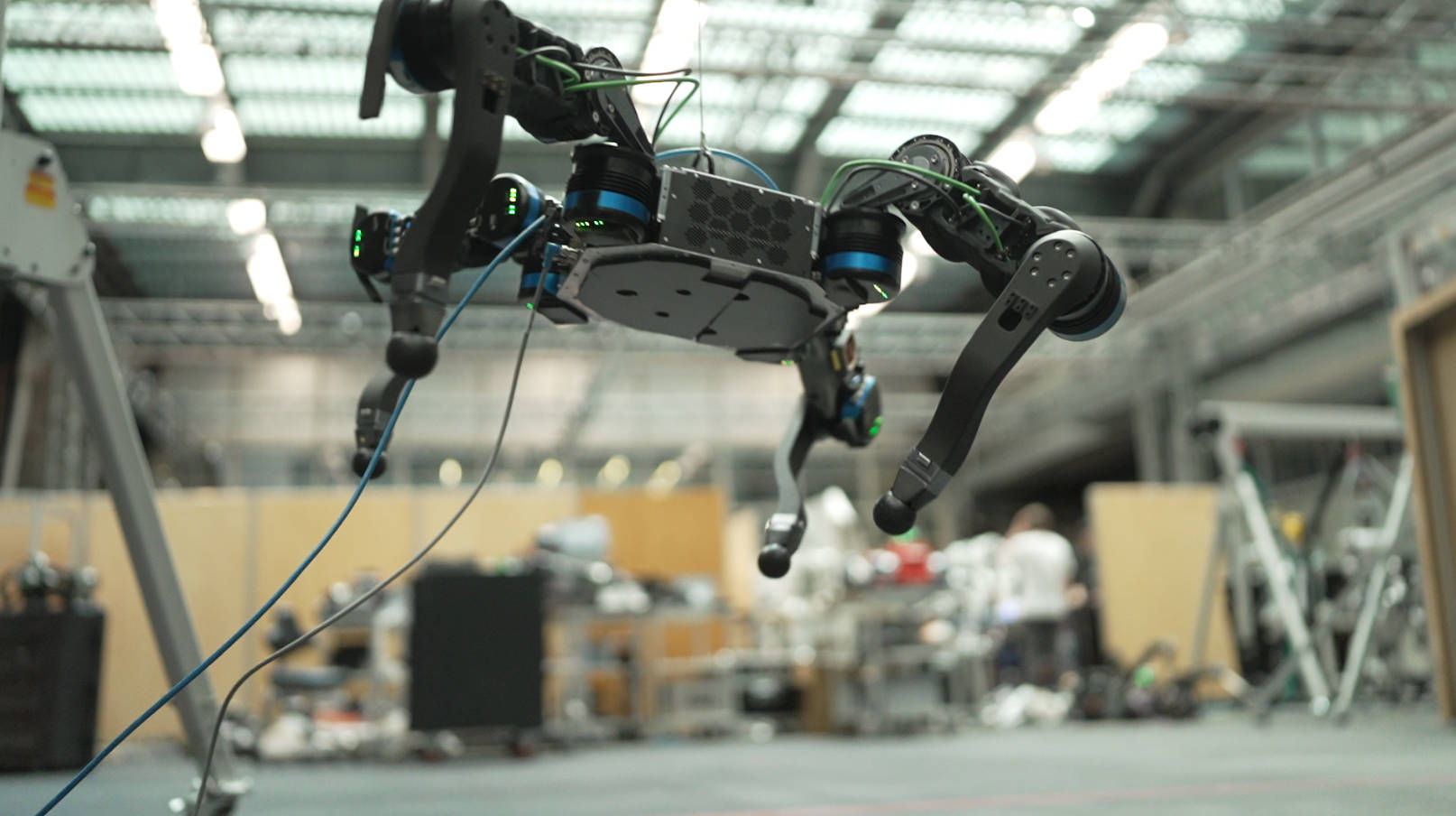}
    \caption{The \textit{Magnecko} robot mid-flight under gravity offloading during hardware deployment of the jumping controller.}
    \label{fig:eyecatcher}
\end{figure}

\begin{figure*}[!t]
    \centering
    \includegraphics[width=\textwidth]{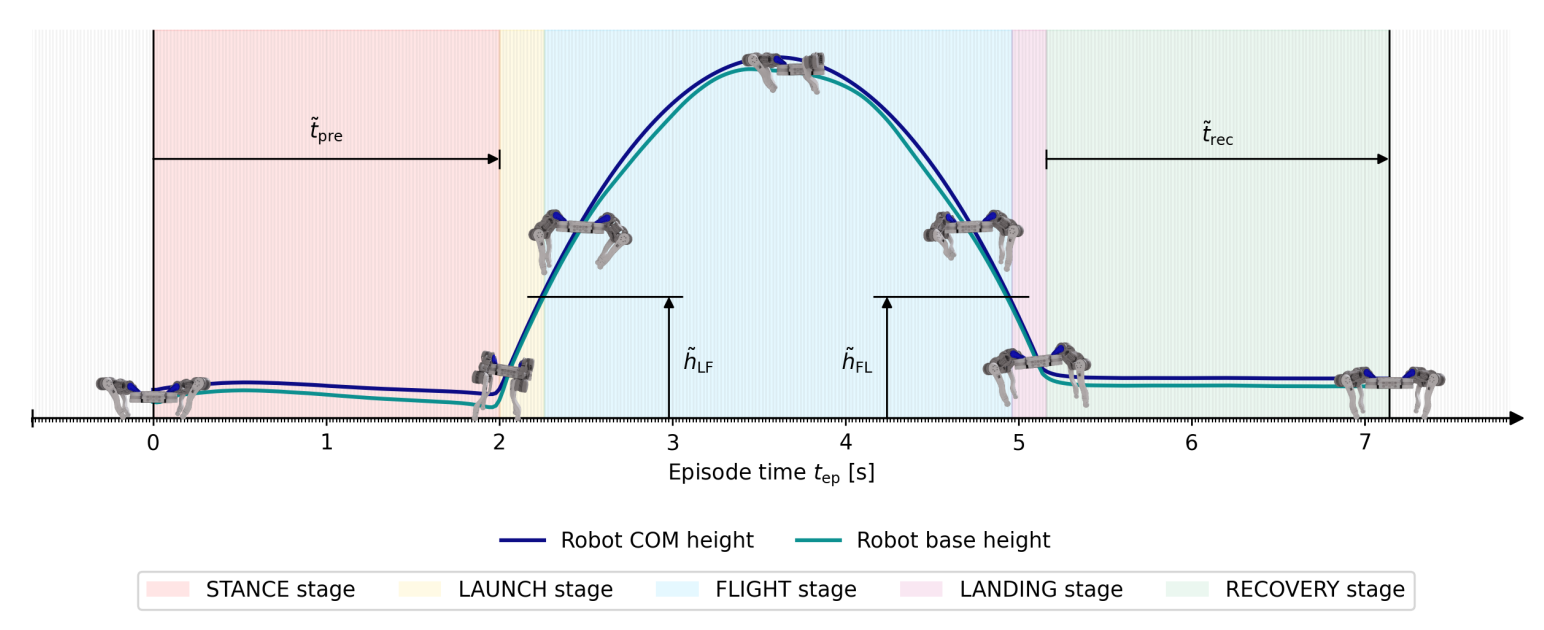}
    \caption{Illustration of the five-stage control scheme used during training and its transition logic. The control policy does not observe the current stage.}
    \label{fig:multistage}
\end{figure*}

Specifically, our contributions are: 
\begin{itemize}
\item We formulate commanded point-to-point jumping in lunar gravity by exploiting the ballistic flight structure to relate the target position and flight trajectory-apex height to the desired launch velocity.
\item We introduce an end-to-end RL framework that learns the full jumping maneuver using a single deployable control policy, while avoiding dense motion references, online trajectory generation, and policy switching.
\item We evaluate the resulting controller on the \textit{Magnecko}\cite{leuthard2024magnecko} quadruped for jumping accuracy, recovery, and robustness to random failures.%as a \textit{LunarLeaper}-relevant surrogate\% 
\item We transfer the resulting controller to gravity offloading hardware to evaluate the sim-to-real profile of the torque and power curves.
\end{itemize}

\section{RELATED WORK}
Quadrupedal jumping has been achieved with trajectory optimization and online MPC tracking~\cite{nguyen2022contact, tanaka2025tethered, nguyen2022continuous}, and with fast online trajectory generation~\cite{yue2024fast, bellegarda2024quadruped}. Chignoli~\textit{et~al.} additionally use an offline jump-feasibility classifier to select omnidirectional jumps in real time~\cite{chignoli2022rapid}. These architectures are highly complex and depend on substantial onboard computation and accurate dynamics models, which makes them ill-suited to a space-deployment setting.

Reinforcement learning (RL), by contrast, is generally more robust and cheaper at rollout time. A common strategy shapes training with a motion reference: Bellegarda~\textit{et~al.} track optimized trajectories with a learned policy~\cite{bellegarda2024robust}, Liu~\textit{et~al.} learn the parameters of a hand-designed foot trajectory for distance-controllable
jumps~\cite{liu2023distance}, and others imitate optimized or example trajectories~\cite{ding2023robust, apostolides2025explosive,
wang2025integrating}. In low gravity, Sang and Wang combine a main-point trajectory generator with staged rewards and shared experience across tasks~\cite{sang2024lunarleap}, which Ye~\textit{et~al.} later augment with
behavior cloning to reach higher relief~\cite{ye2026demoaug}. While motion references improve robustness, they make controller development tedious and limit the jump to the performance of the reference; moreover, reference-based jumping has, to
our knowledge, not been transferred to low-gravity or offloaded hardware.
 
Other works avoid references entirely. Atanassov~\textit{et~al.} obtain a single goal-conditioned policy spanning a wide range of jumps purely through curriculum design, though only under Earth gravity and dense velocity guidance~\cite{atanassov2025curriculum}.
Hierarchical or switching controllers that pair a learned high-level policy with model-based or per-phase low-level controllers add capability but introduce additional modules and potential points of failure at deployment~\cite{yang2023cajun, yang2025agile}. Closest to our setting, Xu~\textit{et~al.} demonstrate continuous lunar pronking on a hardware-in-the-loop digital-twin testbed that pairs an offload with a tilting platform and treadmill~\cite{xu2026dual}. While the latter shows promising results, continuous pronking is undesirable in a mission scenario due to the high cost of transport and unnecessary risk. We instead focus on a single jump, to be deployed only when necessary.

\section{METHOD}
We train and test our controller on an improved version of \textit{Magnecko} \cite{leuthard2024magnecko}. The robot weighs \SI{15.6}{kg}, has a square base with side length \SI{0.25}{m}, and features 3 actuated joints per leg for hip abduction/adduction (HAA), hip flexion/extension (HFE), and knee flexion/extension (KFE). In the default configuration with extended legs, the robot's total extent is a square with side length \SI{0.62}{m}.

We emphasize the deployability of our controller in space. For this reason, we deliberately avoid using secondary estimation modules and recurrent structures in our controller, which could be points of failure in the event of temporary loss of compute.

\subsection{Jumping Physics and Task Formulation}
Fig.~\ref{fig:multistage} illustrates the jumping flight phases over time. 
Due to the lack of an atmosphere on the moon, once the robot has disconnected from the ground, the trajectory of the robot's center of mass is fully determined by the closed-form equations of ballistic flight. A variety of parabolas with varying apex heights connect any two points on the surface. To simplify the learning problem, we uniquely determine the desired parabola path in our command space. The policy receives a command
\begin{equation}\label{eq:command}
    \bm{c} =
    {\left(
        p_{\mathrm{T},x}^{\mathrm{CMD}},
        p_{\mathrm{T},y}^{\mathrm{CMD}},
        p_{\mathrm{A},z}^{\mathrm{CMD}}
    \right)}^\top,
\end{equation} 
where $\bm{p}_{\mathrm{T}}$ is the touchdown position relative to the starting position of the robot under a flat ground assumption, and $p_{\mathrm{A},z}$ is the apex height of the parabola. This allows the user to choose the parabola's shape to suit the desired use case.  The actor's observations are
\begin{equation}
    \bm{o}_a = [\bm{c}, \bm{q}, \bm{\dot{q}}, \bm{a}_{t-1}, \bm{g}_p, \hat{t}_{ep}, \bm{\omega}_{[t,...,t-7]}] \in \mathbb{R}^{67}
\end{equation}
where $\bm{q} \in \mathbb{R}^{12}$ are joint positions, $\bm{\dot{q}} \in \mathbb{R}^{12}$ are joint velocities,  $\bm{a}_{t-1} \in \mathbb{R}^{12}$ are previous actions, $\bm{g}_p \in \mathbb{R}^3$ is projected gravity, $\hat{t}_{ep}$ is time since beginning of the jumping maneuver, and $\bm{\omega} \in \mathbb{R}^{24}$ are the last 8 IMU angular velocity readings. To aid training, the critic additionally receives privileged information, namely $s$ the current stage, $\bm{c}_{\mathrm{f}}\in\mathbb{R}^4\) the foot contact states, $\bm{x}_B \in \mathbb{R}^7$ the world pose of the robot base, $\bm{\xi}_B$, the robot base's twist,  $\bm{p}_{COM} \in \mathbb{R}^3$ the world position of the robot's center of mass, $\bm{v}_{COM} \in \mathbb{R}^3$ the world velocity of the robot's center of mass, and $\bm{v}_L^{CMD} \in \mathbb{R}^3$ the optimal launch velocity given the parabola derived from the command.

The actions $\bm{a} \in \mathbb{R}^{12}$ control the proportional-derivative (PD) controllers on the robot's joints and are updated at $50Hz$. The robot's joint position targets are computed as
\begin{equation}\label{eq:cmd}
    \bm{q}_* = \sigma_a \cdot \bm{a} + \bm{q}_{def}
\end{equation}
where $\sigma_a$ is the action scaling factor, and $\bm{q}_{def} \in \mathbb{R}^{12}$ are the default joint positions.

At each episode, a jump command is sampled according to the ranges in Table~\ref{tab:sim_training_params}. The corresponding launch velocity, an alternative to uniquely define the commanded jumping parabola which we use to reformulate the sparse point-to-point jumping to a launch velocity tracking problem, is reconstructed analytically as
\begin{equation}
    v^{\mathrm{CMD}}_{L,z}=\sqrt{2g_M p_{\mathrm{A},z}^{\mathrm{CMD}}}, \qquad
v^{\mathrm{CMD}}_{L,h}=\frac{g_M d}{2v^{\mathrm{CMD}}_{L,z}}.
\end{equation}

Policies are trained on flat terrain and planar-sloped patches, with terrain parameters, domain randomization, and curriculum settings summarized in Table~\ref{tab:sim_training_params}. Symmetry augmentation is applied over the symmetry axes of the robot, which improves convergence and base-attitude stability without additional simulation \cite{mittal2024symmetry}.

\begin{table}[!ht]
\centering
\caption{Simulation and training parameters.}
\label{tab:sim_training_params}
\footnotesize
\setlength{\tabcolsep}{4pt}
\begin{tabular}{@{}lll@{}}
\toprule
Category & Parameter & Value \\
\midrule
\multirow{5}{*}{\shortstack{Command\\Sampling}}
& Jump distance $d$ & $\mathcal{U}(0,5)\,\mathrm{m}$ \\
& Heading $\phi$ & $\mathcal{U}(-45^\circ,45^\circ)$ \\
& Launch elevation $\vartheta$ & $\geq 45^\circ$ \\
& Launch speed & $\leq 3\,\mathrm{m/s}$ \\
& Apex height $z_A$ & $\geq 0.8\,\mathrm{m}$ \\
\midrule
\multirow{3}{*}{Terrain}
& Flat terrain friction & $\mu_s=\mu_d=1.0$ \\
& Slope inclination & $\mathcal{U}(0^\circ,5^\circ)$ \\
& Slope azimuth & $\mathcal{U}(0^\circ,360^\circ)$ \\
\midrule
\multirow{3}{*}{Randomization}
& Base mass & $\mathcal{U}(-2\,\mathrm{kg},2\,\mathrm{kg})$ \\
& Static friction $\mu_s$ & $\mathcal{U}(0.8,1.2)$ \\
& Dynamic friction $\mu_d$ & $\mathcal{U}(0.6,1.0)$ \\
\midrule
\multirow{3}{*}{Curriculum}
& Initial elevation lower bound & $85^\circ$ \\
& Final elevation lower bound & $45^\circ$ \\
& Annealing schedule & \makecell[l]{$-5^\circ$ every 500 iterations \\ after iteration 1000} \\
\midrule
\multirow{2}{*}{Regularization}
& Actuator-power weight & linearly ramped from $0$ \\
& Ramp duration & first 1000 iterations \\
\bottomrule
\end{tabular}
\end{table}

\begin{table}[!ht]
\centering
\caption{Stage transition conditions and parameters. $b_{TD}$ is set once at least
three feet are simultaneously in contact after takeoff.}
\label{tab:m-trans}
\setlength{\tabcolsep}{4pt}
\begin{tabular}{@{}llcc@{}}
\toprule
Transition & Condition & Param. & Value \\
\midrule
STANCE$\to$LAUNCH   & $(s{=}0)\wedge(t_{ep}\!\ge\!\tilde t_{\mathrm{pre}})$ & $\tilde t_{\mathrm{pre}}$ & $2\,$s \\
LAUNCH$\to$FLIGHT   & $(s{=}1)\wedge(h\!\ge\!\tilde h_{LF})$               & $\tilde h_{LF}$ & $0.75\,$m \\
FLIGHT$\to$LANDING  & $(s{=}2)\wedge(h\!\le\!\tilde h_{FL})$               & $\tilde h_{FL}$ & $0.75\,$m \\
LANDING$\to$RECOVERY& $(s{=}3)\wedge(b_{TD})$                              & $\tilde t_{\mathrm{rec}}$ & $2\,$s \\
\bottomrule
\end{tabular}
\end{table}

\begin{table*}[!ht]
\centering
\caption{Reward terms, expressions, active stages with weights, and parameters.
Stages: STA = STANCE, LAU = LAUNCH, FLI = FLIGHT, LAN = LANDING, REC = RECOVERY; ``ALL'' denotes a term active in every stage. The LAUNCH task rewards $r_{\mathrm{speed}}$ and
$r_{\mathrm{dir}}$, modulated by the temporal kernel $\gamma_t$, are defined in
\eqref{eq:r-speed}--\eqref{eq:launchdiscount}.}
\label{tab:m-rew}
\footnotesize
\setlength{\tabcolsep}{6pt}
\begin{tabular}{@{}llll@{}}
\toprule
Term & Expression & Stage (weight) & Parameters \\
\midrule
\multicolumn{4}{@{}l}{\textit{Task rewards \& penalties}}\\
Start Base Pos.   & $e^{-(e/\sigma)^2},\quad e=\lVert\boldsymbol{p}_{B,xy}-\boldsymbol{p}^{\mathrm{init}}_{B,xy}\rVert$ & STA $0.5$ & $\sigma{=}0.2$ \\
Start Base Head.  & $e^{-(e/\sigma)^2},\quad e=\bar\psi_B-\bar\psi^{\mathrm{init}}_B$ & STA $0.5$ & $\sigma{=}0.7$  \\

Launch Speed      & $\gamma_t\,\big(\lambda~r_{\mathrm{speed}}(\sigma_1) +(1-\lambda)~ r_{\mathrm{speed}}(\sigma_2)\big)$, \eqref{eq:r-speed}~\eqref{eq:launchdiscount} & LAU $25.0$ & $\lambda=0.5, \sigma_1 = 0.6, \sigma_2 =0.08$ \\
Launch Direction  & $\gamma_t\,\big(\lambda~r_{\mathrm{dir}}(\sigma_1) + (1-\lambda)~r_{dir}(\sigma_2) \big)$,~\eqref{eq:r-dir}~\eqref{eq:launchdiscount}     & LAU $15.0$ & $\lambda = 0.3, \sigma_1 = 0.35, \sigma_2 = 0.06$ \\
Bad Launch        & $\mathbb{I}\{c_{NL}\lor c_{BL}\}$~\eqref{eq:term_badlaunch}~\eqref{eq:term_nolaunch}& LAU $-10.0$ & $\tilde{\delta}_{v,mag} =\SI{0.15}{\meter \per \second}$, $\tilde{\delta}_{v,dir} = \SI{15}{\degree}$ \\

Flat Base         & $e^{-\sqrt{\phi_B^2+\theta_B^2}/\sigma}$ & FLI, REC $0.5$ & $\sigma{=}0.5$ \\

Feet in Contact   & $2^{\sum_{l=1}^{4}\mathbb{I}\{c_{f,l}=1\}}/2^{4}$ & STA $2.0$; LAN, REC $0.5$ & --- \\
Recovery Height   & $e^{-(p_{B,z}-p^*_{B,z})^2/\sigma^2}$ & REC $0.2$ & $\sigma{=}0.1,\ p^*_{B,z}{=}0.2\,$m \\
\midrule
\multicolumn{4}{@{}l}{\textit{Regularizations}}\\
Actuator Power    & $\sum_{j=1}^{12}\max(0,\tau_j\dot q_j) +\tfrac{3}{2}R(\tau_j/(k_tG))^2$ & ALL $-5{\times}10^{-5}$ & --- \\
Joint Limits      & $\sum_{j=1}^{12}\max(q_j-q_{\max,j},0)-\min(q_j-q_{\min,j},0)$ & ALL $-2.0$ & --- \\
Base Contact      & $\mathbb{I}\{F_c>\tilde F_c\}$ & ALL $-100.0$ & $\tilde F_c{=}1.0\,$N \\
Leg Contact       & $\mathbb{I}\{F_c>\tilde F_c\}$ & ALL $-20.0$ & $\tilde F_c{=}1.0\,$N \\
\bottomrule
\end{tabular}
\end{table*}

\subsection{Multistage Scheme}

We decompose point-to-point jumping into five distinct stages: STANCE (0), LAUNCH (1), FLIGHT (2), LANDING (3), and RECOVERY (4). All stages share the same regularizations, and each stage has minimal additional rewards to encourage desirable behavior. The active stage is determined by the transition logic shown in Fig.~\ref{fig:multistage} and Tab.\ref{tab:m-trans}. The full reward breakdown and the stage in which each term is active are given in Table~\ref{tab:m-rew}. The current stage is not part of the policy's observations.
 
The STANCE stage gives the robot time to prepare for the jump. We give a reward to prevent deviation from initial position during this stage, but do not impose any type of 'wind-up' reference. After $\tilde t_{\mathrm{pre}}$, the stage is transitioned to LAUNCH.
 
The LAUNCH stage is the most important part of the maneuver. During this stage, the robot receives separate rewards for tracking the launch speed and direction. Both share a convex combination of a wide ($\sigma_1$) and a narrow ($\sigma_2$) kernel: the wide kernel ensures sufficient signal during early training. In contrast, the narrow kernel enforces the precision required once the basic behavior has been discovered. 

The \textit{Launch Speed} reward applies a Gaussian kernel to the relative center of mass (COM) speed error
\begin{equation}
    r_{\mathrm{speed}} = \exp\ \big\{ - \big(
    \frac{\lVert \bm{v}_{COM}\rVert - \lVert\bm{v}_L^{cmd} \rVert}
    {\lVert \bm{v}_L^{cmd} \rVert \cdot \sigma}
    \big)^2
    \big\}
    \label{eq:r-speed},
\end{equation}
where using the relative error makes the term independent of the commanded jump magnitude. We apply the reward twice, once with a large $\sigma$ to encourage discovery of the reward, and once with a smaller $\sigma$ to encourage precision.

The \textit{Launch Direction} reward, 
\begin{equation}
    r_{\mathrm{dir}} = \exp\ \big\{ - \frac{1 - \cos\delta_{\mathrm{dir}}}{\sigma^2} \big\}
    \label{eq:r-dir}, 
\end{equation}
applies an exponential kernel to the cosine alignment error, 
\begin{equation}
    \cos\delta_{\mathrm{dir}} = \frac{\bm{v}_{COM} \cdot \bm{v}_L^{cmd}}
    {\lVert \bm{v}_{COM}\rVert \, \lVert \bm{v}_L^{cmd}\rVert},
    \label{eq:cos-dir}
\end{equation}
where $\delta_{\mathrm{dir}}$ is the angle between the executed and commanded COM launch direction. Using the cosine avoids the noise of explicitly computing the angular error. As with the speed reward, we apply it twice, once with a large $\sigma$ and once with a smaller $\sigma$.

We further emphasize a single, strong jump by applying a temporal exponential kernel 
\begin{equation}\label{eq:launchdiscount}
    \gamma_t       = \exp\{-\rho(t-\tilde t_{\mathrm{pre}})\}\\
\end{equation}
over the task rewards $r_{\mathrm{task}}\in\{r_{\mathrm{speed}},\,r_{\mathrm{dir}}\}$, 
which rewards tracking the required launch velocity instantaneously rather than finding a local minimum, in which the policy makes multiple small hops. Using this trick, the policy learns to perform a 'wind-up' motion at the end of the STANCE stage.

The robot transitions from the LAUNCH stage to FLIGHT once a minimum height $\tilde h_{LF}$ is reached. We use two 'bad launch' termination conditions. The first is triggered if the robot has failed to launch completely after a set time $\tilde{t}_{L,max}$
\begin{equation}\label{eq:term_nolaunch}
    c_{NL} = (s=LAUNCH) \land (t_{ep} - \tilde{t}_{pre} > \tilde{t}_{L,max} )
\end{equation}
where $\tilde{t}_{L,max} = \SI{2}{\second}$ for this work. The second termination condition is triggered when the robot successfully launches, but does so with unacceptable launch parameters:
\begin{equation}\label{eq:term_badlaunch}
\begin{aligned}
    c_{BL} = \big((\delta_{v,mag} > \tilde{\delta}_{v,mag}) \lor 
    (\delta_{v,dir} > \tilde{\delta}_{v,dir}) \big) \\
    \land (t_{ep} - \tilde{t}_{pre} > \tilde{t}_{L,max} )
\end{aligned}
\end{equation}
where $\delta_{v,mag}$, $\delta_{v,dir}$ are launch velocity and direction tracking error respectively. The robot also incurs a large penalty when terminated under either of these conditions. This both accelerates and stabilizes the learning, as the robot spends less time in junk states where it failed to launch properly. Importantly, this training setup causes the robot to launch immediately at $\hat{t}_{ep} \approx \tilde{t}_{\mathrm{pre}}$ with $\bm{v}_L \approx \bm{v}_L^{CMD}$. Notably, due to the precisely timed and executed launch velocity and as we provide a timer observation, the robot can infer its progress on the flight trajectory, without dedicated exteroceptive sensing information. In fact, in conjunction with the asymmetric actor-critic setup, and the type of stage transition heuristics in \cref{tab:m-trans} on which the set of active rewards ultimately depends, this not only allows us to synthesize the distinct stage-specific control behavior within a single control policy but to do so without relying on dense motion references, neither during training nor deployment, and with minimal mission-compatible sensing requirements.

During the underactuated FLIGHT stage, the \textit{Flat Base} reward incentivizes the robot to leverage conservation of angular momentum and coordinated limb motion for in-flight base tilt $\delta_{\mathrm{Tilt}}$ control, resp. to keep the base stable.

The stage transitions to LANDING when the robot's base falls below the threshold $\tilde{h}_{FL}$. During the LANDING stage, the robot is rewarded for maximizing the number of feet in contact with the ground, to encourage spreading out the impact as much as possible.

As soon as at least 3 feet are in contact with the ground, the stage transitions to the terminal RECOVERY stage. In this stage, the robot is rewarded for staying put and adopting a stable stance. This allows for easy transition back into a 'standard' locomotion control.

\subsection{Simulation and Hardware Setup}
We train our control policy in \textit{IsaacLab}\cite{mittal2025isaac} with the Proximal Policy Optimization (PPO) algorithm\cite{schulman2017proximal} and the hyperparameters from \textit{Rudin et al.}\cite{rudin2022learning}. We use multilayer perceptrons with hidden layer sizes of \{512, 256, 128\} and \textit{ELU} activation functions for both the actor and critic networks. The simulation joint parameters of the robot were identified using PACE\cite{bjelonic2025towards}. The control policies were trained for $10{,}000$ apex height, on an RTX4090 GPU.

For hardware deployment, \textit{Magnecko} was attached to a custom gravity offloading mechanism that approximates lunar-gravity loading. The mechanism uses a constant-force spring reel to provide an upward tether force that offsets approximately 5/6 of the robot's gravitational load, leaving an effective weight close to that under lunar gravity. The legs are provided feedforward gravity offloading torques as in \textit{Arm et al.}\cite{arm2025efficient} such that the legs experience the same gravity as in the simulation. Although the setup does not reproduce fully unconstrained lunar flight, it provides sufficient fidelity and a practical hardware testbed for vertical jumping experiments and for comparing simulation and hardware results.

\section{RESULTS}

\subsection{Jumping Accuracy}
\label{ss:jumping_acc}

\begin{figure}[!t]
    \centering
    \includegraphics[width=\columnwidth]{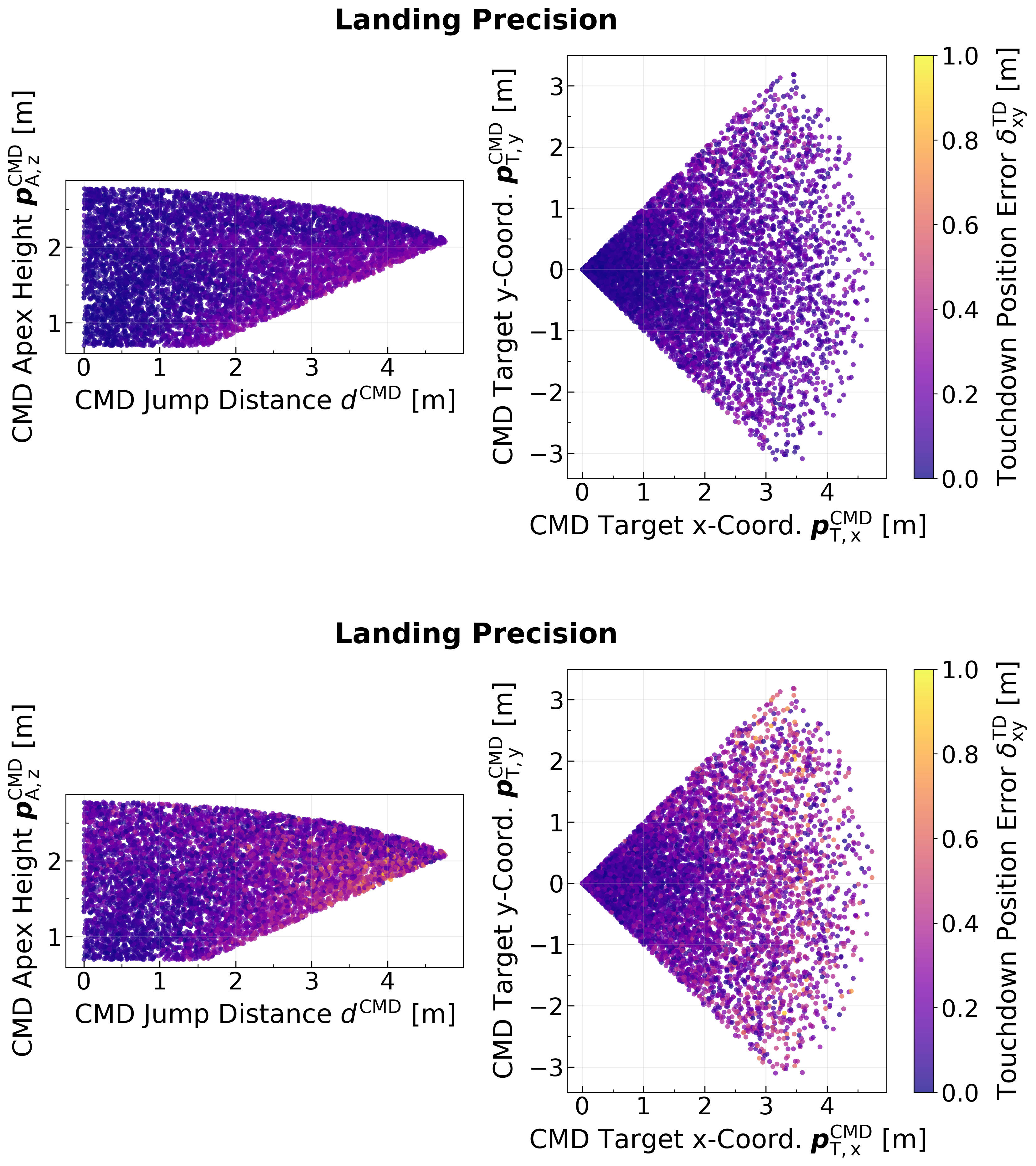}
    \caption{Landing accuracies corresponding to $10{,}000$ commands covering the full command workspace. Top: Flat Terrain. Bottom: Sloped Terrain.}
    \label{fig:landing_prec_rough}
\end{figure}

We evaluated the landing position accuracy over the trained workspace for both flat and sloped terrain. We randomly sampled $10{,}000$ triplets of landing position $x$, and landing position $y$ and apex height. These commands were then rolled out once on flat terrain and once on sloped terrain. The sloped terrain's incline was sampled from $\mathcal{U}(\SI{0}{\degree},~\SI{5}{\degree})$ and its direction was sampled from $\mathcal{U}(\SI{0}{\degree},~\SI{360}{\degree})$. The landing accuracy is shown in \cref{fig:landing_prec_rough}. In both scenarios, tracking accuracy appears to be better when the commanded apex is larger and landing accuracy degrades with commanded jumping distance. The degradation is more pronounced for the sloped scenario.

\begin{figure}[!t]
    \centering
    \includegraphics[width=\columnwidth]{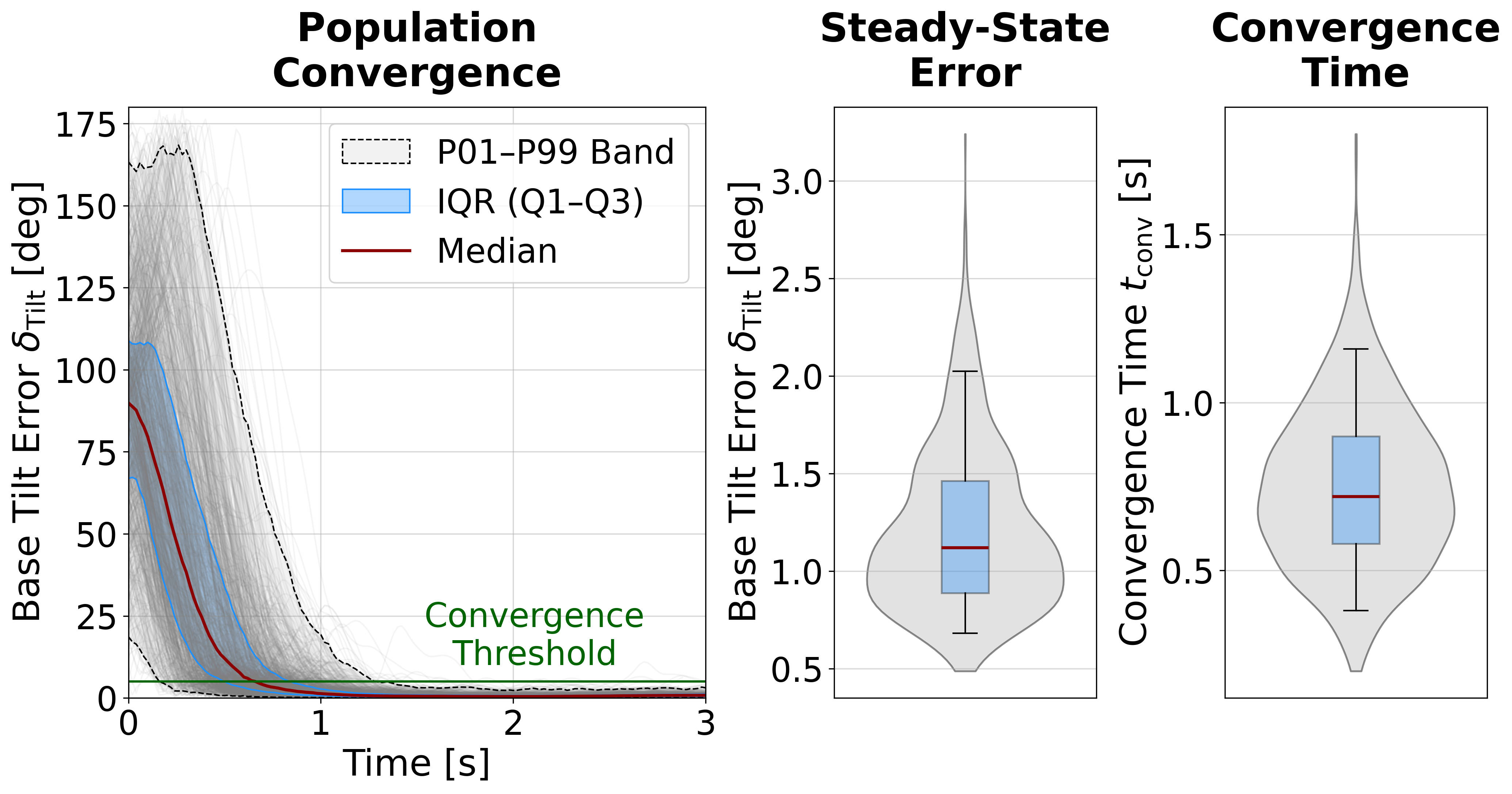}
    \caption{Mid-air recovery behavior of the controller.}
    \label{fig:air_recovery}
\end{figure}

\subsection{Robustness}
\label{ss:robustness}

\begin{table}[!t]
\centering
\begin{tabular}{ccc}
\textbf{Failure duration} & \textbf{Flat Crashes} & \textbf{Sloped Crashes} \\ \hline
\SI{0.0}{\second} (nominal) & 7.5\% & 11.9\%          \\
\SI{0.2}{\second}   & 7.8\%         & 12.5\%          \\
% \rowcolor[HTML]{EFEFEF}
\SI{0.5}{\second}   & 11.3\%        & 15.6\%          \\
\SI{1.0}{\second}   & 16.7\%        & 20.3\%         
\end{tabular}
\caption{Failure rates under fixed-duration control freeze.}
\label{tab:failure-rate}
\end{table}

\begin{figure}[!t]
    \centering
    \includegraphics[width=\columnwidth]{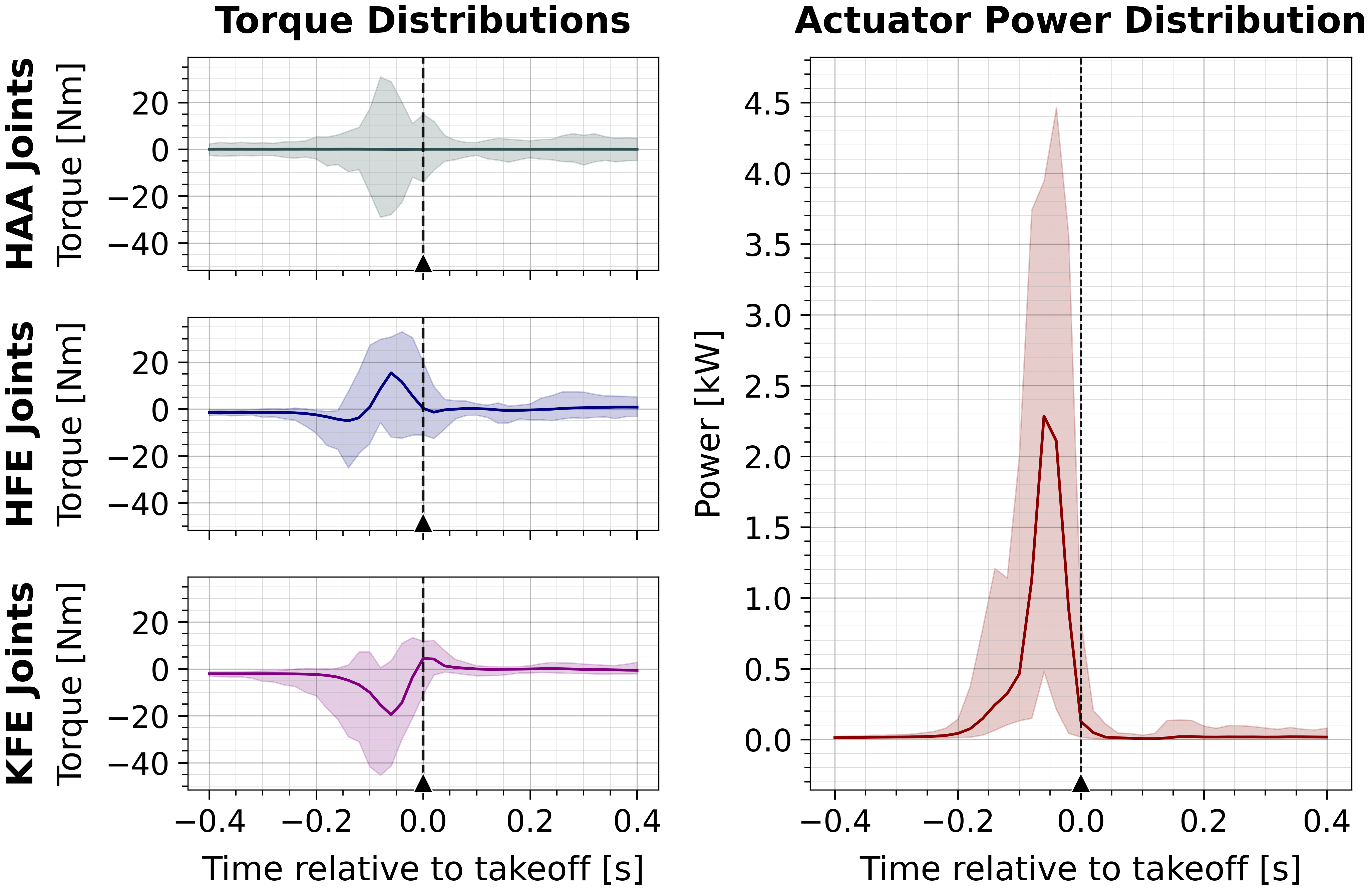}
    \caption{Simulated torque and power demands of the robot around takeoff; dashed line and black triangle mark takeoff time.}
    \label{fig:torquepowerdemands}
\end{figure}

\begin{figure*}[!t]
    \centering
    \includegraphics[width=\textwidth]{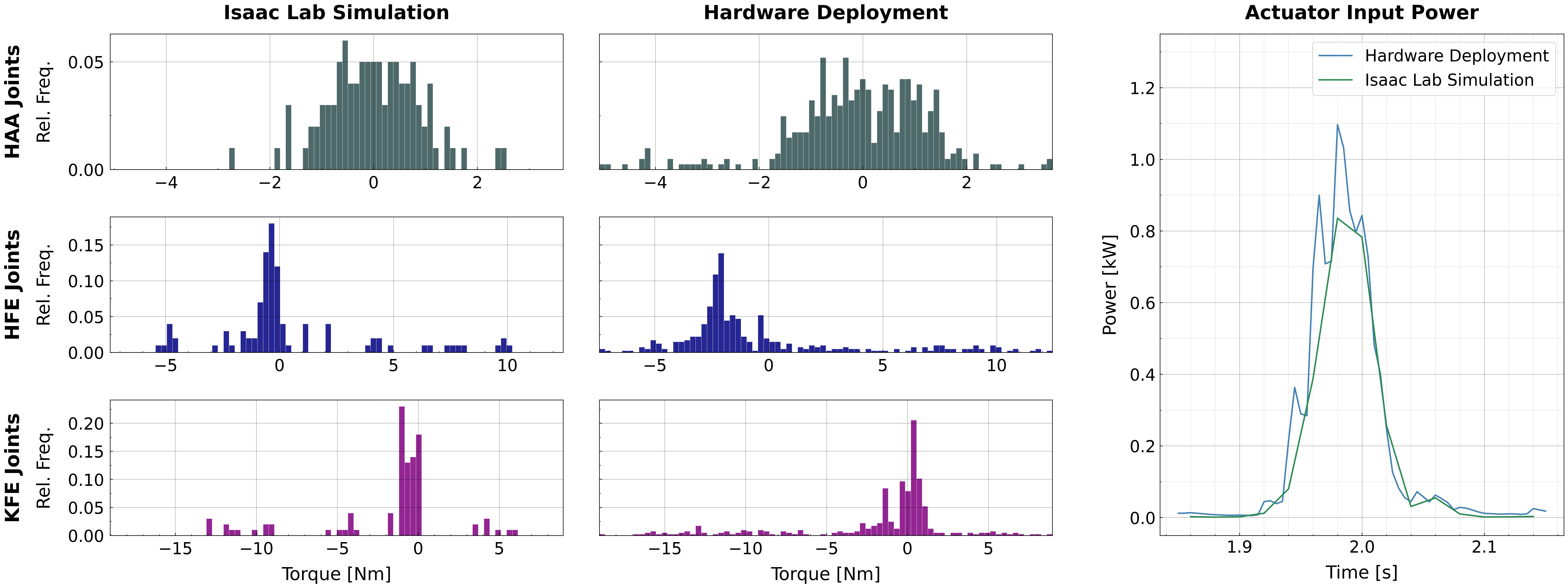}
    \caption{Sim-to-real joint value transfer. Left: per joint group torque distributions during the launch for simulation and hardware deployment. Right: total actuator input power demand during the launch for simulation and hardware deployment.}
    \label{fig:sim2real}
\end{figure*}

\begin{figure}[!t]
    \centering
    \includegraphics[width=\columnwidth]{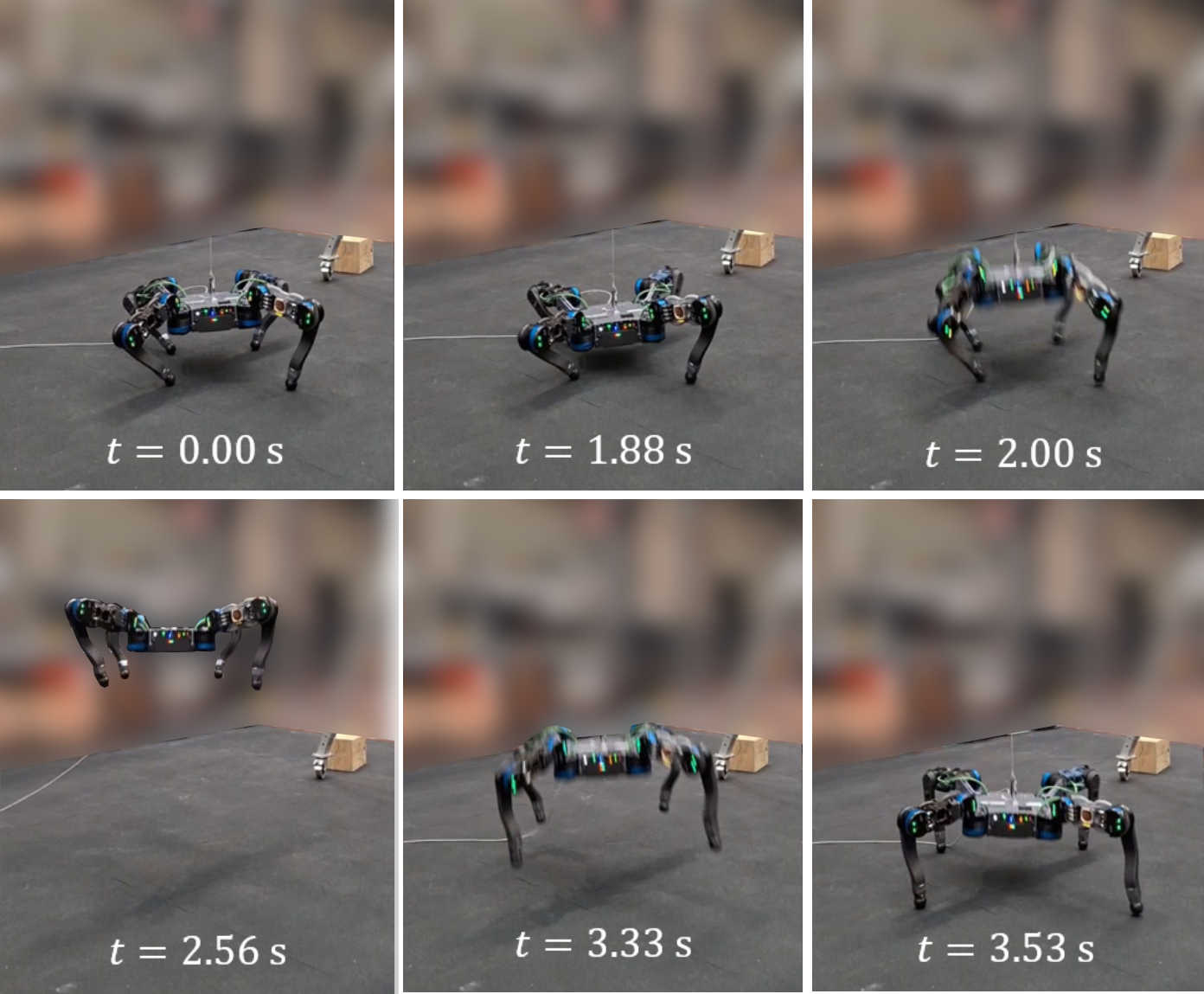}
    \caption{Example of the jumping sequence deployed on \textit{Magnecko}.}
    \label{fig:HWjump}
\end{figure}

We investigated the robustness of the trained control policy in two experiments. In the first experiment, we initialized the control policy mid-flight in a random orientation, as one might encounter, for example, as a result of a single-event upset and consequent forced reboot of the controlling computer. The starting configuration roll, pitch, and yaw angles were sampled from $\mathcal{U}(-\pi~\SI{}{\radian},\pi~\SI{}{\radian})$, and starting angular velocities were sampled from $\mathcal{U}(-\pi~\SI{}{\radian \per \second}, \pi~\SI{}{\radian \per \second})$. We then measured the time required for the control policy to
reorient the robot base to its nominal tilt. The convergence threshold chosen for this was \SI{5}{\degree}. The median convergence time was \SI{0.7}{s}, with \SI{99}{\%} converging in under \SI{1.5}{s}. A representative mid-air recovery sequence is shown in \cref{fig:air_recovery}.

In the second experiment, we ran $10{,}000$ rollouts on flat and sloped terrain using the same randomly sampled commands as in \cref{ss:jumping_acc}. We then logged contact forces between the robot base and the terrain as a proxy for landing robustness. Any contact with the robot's base and the ground was considered as a crash regardless of contact force magnitude. On flat terrain, \SI{92.5}{\%} of robots completed the jump without any contact forces. For the robots jumping on sloped terrain, \SI{88.1}{\%} completed the jump without any contact forces. We then reran the experiment with simulated failures. For each rollout, we briefly disabled the control policy's target inputs to the joints at a random time during the episode, simulating a compute failure. Upon regaining control, we flushed the IMU angular velocity history buffer to simulate loss of previous readings during compute shutdown. We performed the experiment three times, simulating loss of control for $[0.2, 0.5, 1.0]$ seconds. The joints themselves remained active with the last sent PD target, as we assume they have an inbuilt driver. \cref{tab:failure-rate} shows the failure rates.

\subsection{Torque and Power Demands}
\label{ss:torque_power_demands}
To characterize the mechanical and electrical demands imposed by jumping, both joint-level torques and actuator input power demands were analyzed. In this regard, the takeoff sequence is the dominant event, with pronounced torque and power peaks immediately before takeoff is recorded. In a separate experiment on sloped terrain, we summarized the mean and envelope of the torques for each joint group, and the combined input power demands over all actuators around takeoff, indicated by the black triangles in \cref{fig:torquepowerdemands} for a command set that systematically covers the entire trained command domain. The \textit{Magnecko} actuators have a peak torque of \SI{56}{Nm}.

\begin{comment}
Additionally, we obtain a quantitative indication of the energetic efficiency of the learned jumping behavior, by evaluating the \textit{cost of transport} (CoT). To assess how efficiency varies with jump distance, the CoT is evaluated for jump distances from \SI{1}{m} to \SI{5}{m}, always for the an admissible launch elevation angle as close to \SI{45}{\degree} as possible. The resulting CoT values are then averaged over distance intervals as summarized in \cref{tab:cot}. The results indicate a decreasing CoT with increasing jump distance.

\begin{table}[H]
    \centering
    \caption{Average CoT over different jump-distance ranges.}
    \label{tab:cot}
    \begin{tabular}{lcccc}
        \toprule
        Distance range [m] & $[1,2]$ & $[2,3]$ & $[3,4]$ & $[4,5]$ \\
        \midrule
        Average CoT & $3.2$ & $2.9$ & $2.6$ & $2.5$ \\
        \bottomrule
    \end{tabular}
\label{tab:cot}
\end{table}
\end{comment}

\subsection{Hardware Feasibility}

We investigated the sim-to-real transfer behavior. The robot was attached to the gravity offloading device described in \cite{arm2025efficient}. As the offload was incapable of translational movements, we commanded purely vertical jumps with an apex height of \SI{0.8}{\meter}, and $x$ and $y$ positional commands of \SI{0}{\meter}, effectively jumping in place. The height of \SI{0.8}{\meter} was chosen due to the limits of the gravity offload. We focus on takeoff, as it is the most critical part of the maneuver, since the trajectory is fully determined at takeoff in the absence of an atmosphere. \cref{fig:sim2real} compares the torque distributions measured in the hardware experiments against a matched simulation rollout. It also reports the total actuation hardware power usage and the corresponding simulation estimate around takeoff. \cref{fig:HWjump} demonstrates the hardware jumping sequence.

\section{DISCUSSION}
% The magnitude of the accuracy error generally remains less than $0.3$ m when using launch trajectories with elevation angle $\geq \SI{60}{\degree}$. This error is smaller than the robot's extent (\SI{0.62}{\meter}). 
% %, which measures $0.62m$ by $0.62m$ in default configuration. 
% We deem this accuracy acceptable for mission scenarios, as the main purpose of the controller is to overcome otherwise untraversable terrain rather than require pinpoint accuracy. The accuracy when jumping on slopes is degraded, in part because our ballistic assumption holds only for flat ground. The operator or high-level planner can control and adjust the parabola parameters to improve landing accuracy on irregular terrain.

For launch trajectories with elevation angles $\geq \SI{60}{\degree}$, the landing error generally remains below \SI{0.3}{\meter}, less than half of the robot's \SI{0.62}{\meter} total extent. This indicates that the controller provides sufficient landing accuracy for its intended role as a contingency maneuver to overcome otherwise untraversable terrain, where exact touchdown placement is less critical than safely reaching the target region. The reduced accuracy on sloped terrain is consistent with the flat-ground assumption used to derive the commanded ballistic trajectory. This limitation can be mitigated during mission operation by adjusting the commanded touchdown position and apex height, either manually by an operator or automatically by a higher-level planner.

% The robustness experiments demonstrated that the robot can quickly and consistently reorient in mid-air under the proposed controller. We showed that the control policy successfully completed collision-free landings in $>88\%$ of rollouts on both flat and sloped terrain. When applying a random compute failure of up to \SI{1}{\second} during the maneuver, the control policy still survives $79.7\%$ of the time on sloped terrain. The maneuver could be risky for a real mission with such possible substantial compute failure, but it would be a plausible option in cases where the mission could not otherwise proceed.

The robustness experiments demonstrate that the proposed controller can quickly and consistently reorient the robot in mid-air. In nominal rollouts, the policy achieved collision-free landings in $> \SI{88}{\%}$ of trials on both flat and sloped terrain. Importantly, the controller remained highly resilient even under severe compute interruptions, such as a random control freeze lasting up to \SI{1}{\second}, and the robot still survived $\SI{79.7}{\%}$ of sloped-terrain rollouts. This robustness to temporary loss of control authority is encouraging for deployment-constrained systems, where transient faults may occur. Although mission use would require appropriate risk assessment, the results support jumping as a credible contingency behavior for escaping otherwise untraversable terrain.

The torque and power evaluation confirms the running hypothesis that actuation requirements are mostly determined by the takeoff sequence, which imposes demanding short-time torques and powers, provides a broad reference for system design, and shows that even for the most demanding jumps, demands respect \textit{Magnecko}'s hardware constraints. 

\begin{comment}
The CoT evaluation suggests that, for longer jumps, the energetic cost associated with launch is increasingly compensated by the prolonged ballistic flight phase, during which comparatively little energy is consumed. Nevertheless, when compared to a lunar walking policy developed for the same robot platform, which achieves CoT values in the range of approximately 0.6 to 1.3, the jumping policy remains noticeably less efficient over the distance range considered in this work.
\end{comment}

The hardware deployment shows that the sim-to-real gap can be bridged for vertical real-world jumping. Horizontal jumping could not be validated due to limitations in the gravity offload.

\section{CONCLUSION AND FUTURE WORK}
We have introduced a method for robust point-to-point jumping for a legged robot in lunar gravity. The method is deliberately simple in architecture to facilitate deployment in space applications where secondary estimation modules and recurrence may introduce deployment risk. We have shown that jumping is feasible to implement on hardware and robust enough to deploy as a last resort should one encounter terrain that would be otherwise untraversable.
While the validation done with the gravity offload provides a promising hardware transfer, due to limitations in the offload's design we were unable to validate horizontal jumps. The most immediate followup to this work should be developing and validating on an improved gravity offloading device which displays low inertia in the horizontal direction to conclusively prove hardware transfer fully possible. Additionally, the method's robustness and success rate could be further improved to make it useful beyond deployment as a last resort. Future work should aim to improve robustness while retaining a simple architecture. Finally, the ground contact is currently modeled as rigid, whereas the lunar regolith provides a soft contact that should be properly modeled in future work.

\section*{ACKNOWLEDGMENTS}
We thank the \textit{Magnecko} team for their continued support with repairing the robot, the maintenance of the robot's software and hardware, and, more generally, for building a well-engineered and easy-to-use system. \\
LunarLeaper acknowledges funding through the CHiwi Foundation, SERI's NASO program (1-012571, LunarLeaper 2024), Moog Inc., maxon, and MARVIS.

\bibliographystyle{IEEEtran}
\bibliography{references_lunarjumping, references}
\end{document}